\documentclass[letterpaper, 10pt, conference]{Docs/ieeeconf}

\usepackage{times}
\usepackage{tabularx}
\usepackage{subcaption}
\usepackage{multirow,multicol, array}
\usepackage[font={small}]{caption}   
\usepackage{graphicx}
\usepackage{dblfloatfix}
\usepackage{wrapfig}
\usepackage{siunitx}
\usepackage{soul}

\let\labelindent\relax
\usepackage{enumitem}

\usepackage{amsmath,amsthm,amssymb,amsfonts}
\usepackage[titlenumbered,ruled]{algorithm2e}
\usepackage{textcomp}
\usepackage{acronym}
\usepackage{balance}
\usepackage{mdwmath}
\usepackage{bm}
\usepackage{mdwtab}
\usepackage{array}
\usepackage[usenames,dvipsnames]{color}
\usepackage{eqparbox}
\usepackage{cite}

\usepackage{color}
\usepackage{psfrag}
\usepackage{epsfig}
\usepackage{url}

\usepackage{epstopdf}
\usepackage{booktabs}
\usepackage{blindtext}

\usepackage[colorinlistoftodos,prependcaption,textsize=tiny]{todonotes}

\renewcommand{\emph}{\textit}

\newtheorem*{lemma*}{Lemma}

\newtheorem*{problem*}{Problem}

\makeatletter
\newcommand\fs@spaceruled{\def\@fs@cfont{\bfseries}\let\@fs@capt\floatc@ruled
    \def\@fs@pre{\vspace{5\baselineskip}\hrule height.8pt depth0pt \kern2pt}%
    \def\@fs@post{\kern2pt\hrule\relax}%
    \def\@fs@mid{\kern2pt\hrule\kern2pt}%
    \let\@fs@iftopcapt\iftrue}
\makeatother

\IEEEoverridecommandlockouts
\graphicspath{{images/}}

\begin{document}

	
\title{Development and Testing of a Novel Automated Insect Capture Module for Sample Collection and Transfer}
	
\author{Keran Ye, Gustavo J. Correa, Tom Guda, Hanzhe Teng, Anandasankar Ray, and Konstantinos Karydis
	\thanks{Keran Ye, Gustavo J. Correa, Hanzhe Teng, and Konstantinos Karydis are with the Dept. of Electrical and Computer Engineering, University of California, Riverside. Email: \{kye007, gcorr003, hteng007, karydis\}@ucr.edu. Tom Guda and Anandasankar Ray are with the Dept. of Molecular, Cell and Systems Biology, University of California, Riverside. Email: \{tom.guda, anandr\}@ucr.edu.}
	\thanks{We gratefully acknowledge the support of DARPA under grant \# HR0011835180, NSF under grants \# IIS-1724341 and \# IIS-1901379, and ONR under grant \# N00014-18-1-2252. Any opinions, findings, and conclusions or recommendations expressed in this material are those of the authors and do not necessarily reflect the views of the funding agencies.}
	}
	
\maketitle
\thispagestyle{empty}


\begin{abstract}
There exists an urgent need for efficient tools in disease surveillance to help model and predict the spread of disease. The transmission of insect-borne diseases poses a serious concern to public health officials and the medical and research community at large. In the modeling of this spread, we face bottlenecks in (1) the frequency at which we are able to sample insect vectors in environments that are prone to propagating disease, (2) manual labor needed to set up and retrieve surveillance devices like traps, and (3) the return time in analyzing insect samples and determining if an infectious disease is spreading in a region. To help address these bottlenecks, we present in this paper the design, fabrication, and testing of a novel automated insect capture module (ICM) or trap that aims to improve the rate of transferring samples collected from the environment via aerial robots. The ICM features an ultraviolet light attractant, passive capture mechanism, panels which can open and close for access to insects, and a small onboard computer for automated operation and data logging. At the same time, the ICM is designed to be accessible; it is small-scale, lightweight and low-cost, and can be integrated with commercially available aerial robots. Indoor and outdoor experimentation validates ICM's feasibility in insect capturing and safe transportation. The device can help bring us one step closer toward achieving fully autonomous and scalable epidemiology by leveraging autonomous robots technology to aid the medical and research community.
%
\end{abstract}

\section{Introduction}
Autonomous mobile robots can play a significant role in applications that require sample collection, handling, transfer, and testing~\cite{Yangeabb5589,tirado2019drones,delmerico2019current}.  
Robotic assistance may speed up such processes, especially at large spatio-temporal scales\cite{delmerico2019current,das2011towards,qian2017ground,howard2006experiments}, and reduce potential risks to human personnel tasked to collect and handle samples as well as mitigate lack of sufficient numbers of trained personnel to perform such tasks. 
Enabling rapid and large-scale sample collection and analysis might in turn help model and predict the spread of infectious disease in people, animals, and plants faster and more accurately~\cite{tirado2019drones}.

The spread of arthropod-borne diseases is among the most serious concerns faced by public health officials and the medical community at large~\cite{sarwar2015insect}. Vector-borne diseases are infectious diseases or illnesses transmitted through insects such as mosquitoes, sand flies, ticks, fleas, lice, bugs and flies \cite{sarwar2015insect}.  
For example, the housefly (\emph{Musca domestica})---which is the main insect motivating the work herein---is known to carry pathogens that can cause serious diseases in humans and animals. Numerous pathogens including bacteria, viruses, fungi and parasites have been associated with the insect~\cite{khamesipour2018systematic}. Houseflies can carry and transmit diseases between vertebrate hosts without amplification or development of the organism within the vector~\cite{sarwar2015insect}.  
The ability to sample insect populations in an area at large, such as the housefly, and rapidly transport them back to the lab for analysis to investigate if they may carry pathogens, can serve as an early indicator that the population in the vicinity of sampling points might be contracting pathogens and signal that prevention measures must be taken~\cite{tirado2019drones,koch2015mapping}.

\begin{figure}[!t]
	\vspace{6pt}
	\centering
\includegraphics[trim={5cm 1.25cm 5cm 2.8cm},clip,width=0.85\linewidth]{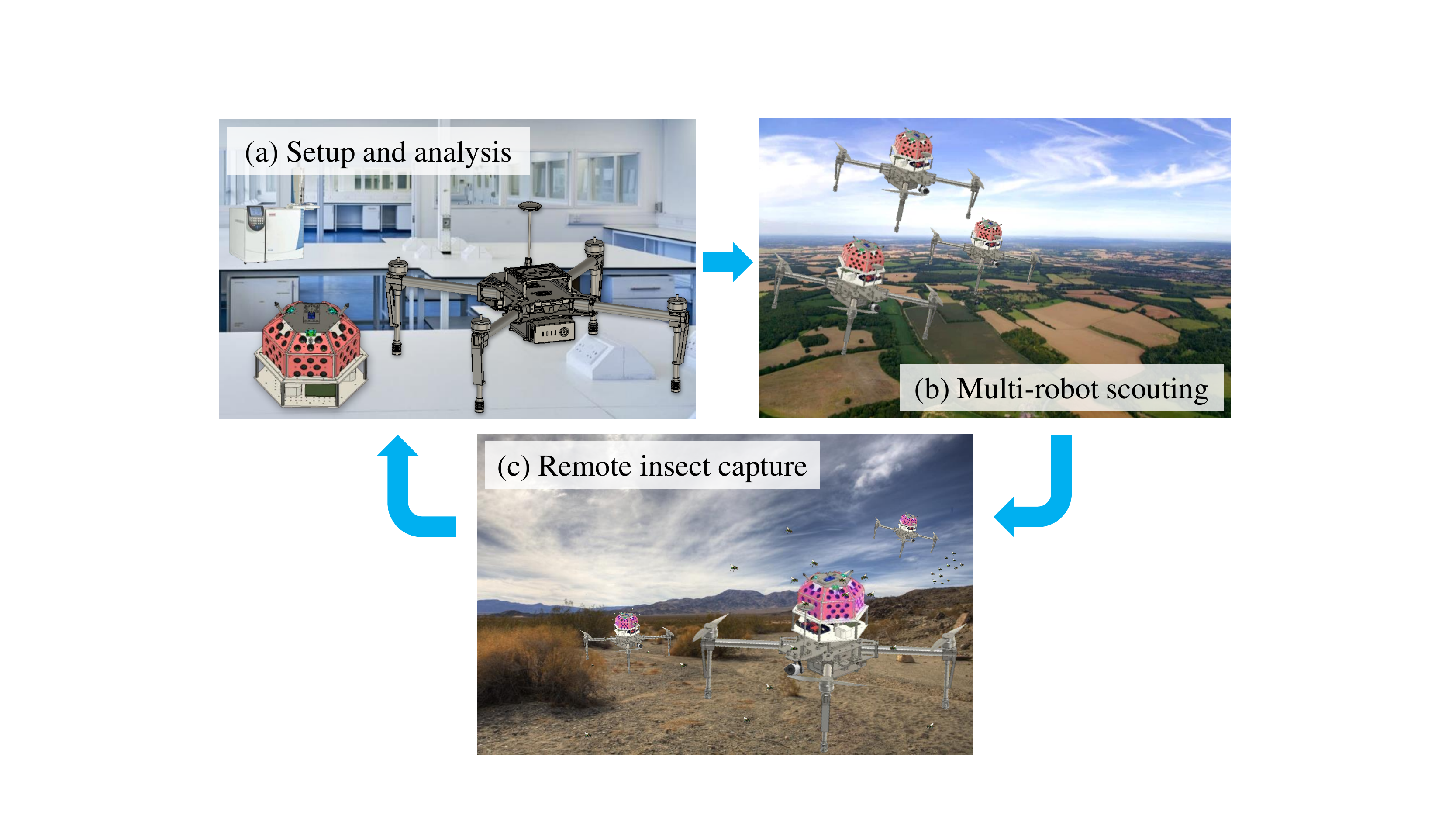}
	\vspace{-3pt}
	\caption{Our overarching vision on remote sample collection and transfer using autonomous robots. (a) Remote sampling mechanism preparation and post-analysis. (b) Robot team scouting for sampling locations. (c) Remote insect capture. This paper focuses on introducing a new insect capture module and testing its feasibility on a single-robot proof-of-concept paradigm.}
	\label{fig:end2end}
	\vspace{-18pt}
\end{figure}

Our overarching goal is to investigate how autonomous mobile robots can serve as the means to facilitate sample (in this case, insects) collection and transfer (Fig.~\ref{fig:end2end}). 
Robots can be deployed to specific locations where insect sampling will be performed. Once the robots arrive at their destination, the sampling mechanism switches on for a desired amount of time to capture insects in the vicinity.  The robots are geo-tagged and eventually return back to the lab, carrying captured insects. Once back to the lab, analysis of the insects captured (such as identification of species, or DNA/RNA profiling) can reveal vital information necessary for surveillance. The process can then repeat at a finer scale to improve the resolution and confidence in those possible areas of interest.

\begin{figure}[!t]
	\vspace{4pt}
	\centering
\includegraphics[trim={0cm 0cm 14cm 9cm},clip,width=0.85\linewidth]{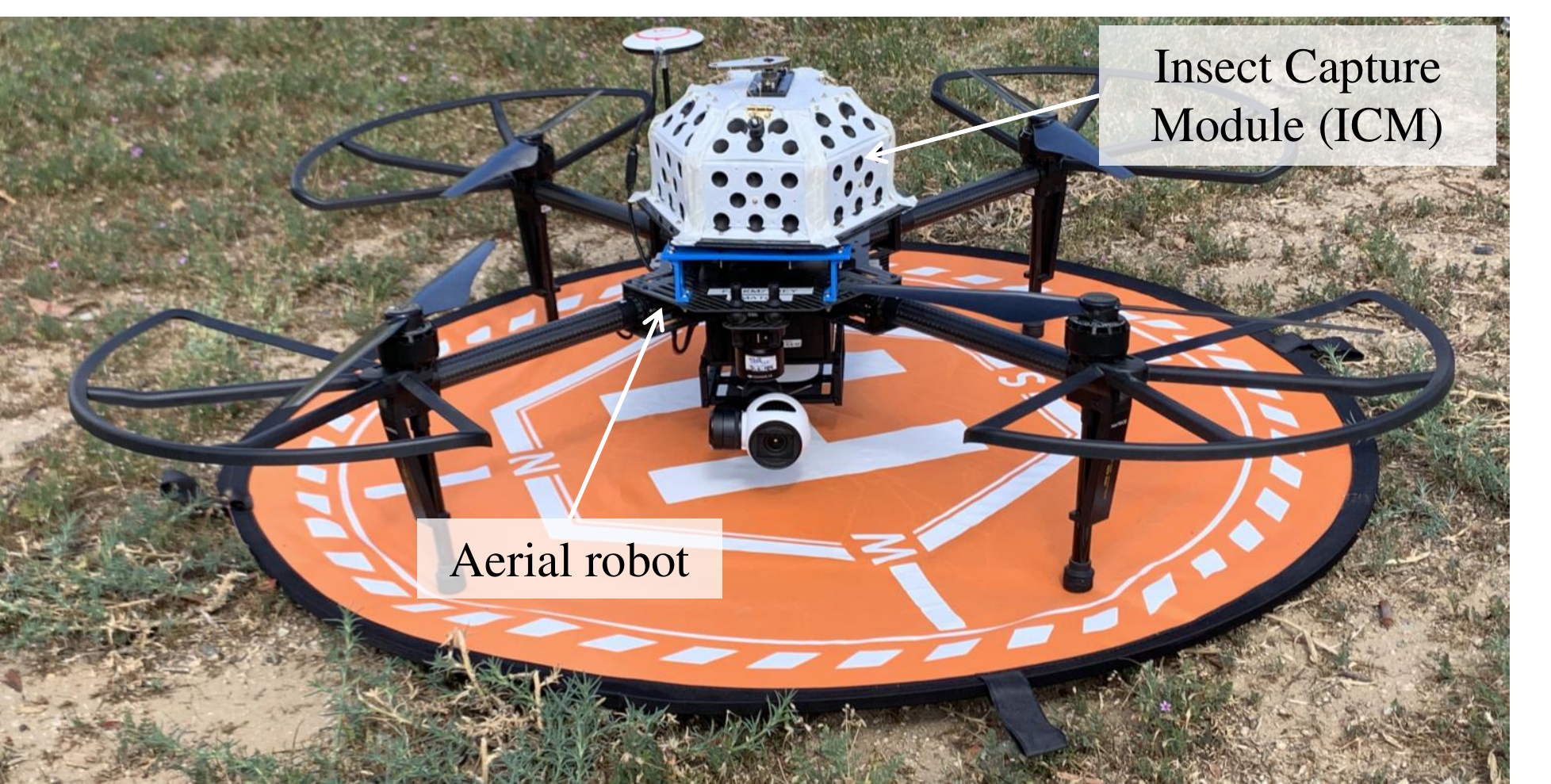}
	\vspace{0pt}
	\caption{The novel insect capture module (ICM) developed in this work while mounted on a commercially-available aerial robot.}
	\label{fig:flyability}
	\vspace{-15pt}
\end{figure}

{\bf This paper, specifically, focuses on the design, feasibility analysis and testing of a novel automated insect capture module (ICM) that can get airborne on a commercially-available aerial robot (Fig.~\ref{fig:flyability}).} 
We employ aerial robots because they can rapidly travel to remote areas that may be otherwise inaccessible. Out of all types of aerial robots, quadcopters offer promise in remote sampling because of their intuitive control, high maneuverability in confined environments, reconfigurability, and comparatively lower costs\cite{karydis2017energetics}. 
Past literature on insect traps have demonstrated different types of effective lures for attracting and capturing insects. Typical insect attractants include odorous food~\cite{pinniger1990food}, chemical odorant baits~\cite{mcneil1991behavioral}, and ultraviolet (UV) light~\cite{barghini2012uv}. Once an insect is attracted close to the trap, entrapping the insect is traditionally achieved with water~\cite{schneider1982flying}, fans~\cite{santos2018fan, miller2001counterflow}, or adhesive surfaces~\cite{shanahan1977baseboard}.

Existing insect traps are static and have no automation \cite{linn2016building} therefore presenting an opportunity for development. A notable exception is the Microsoft Research Premonition project~\cite{dacko2020mosquito, ching2017insect}. The Premonition trap seeks to selectively capture desired mosquito species by using automated analysis of wing beat frequency of individual mosquitoes before capture. However, one limitation of this effort is that the traps at their current form cannot be carried by smaller aerial robots.  Hence, it may be technically challenging and prohibitively expensive to deploy multiple of these devices in instances where scaling up numbers are important to enable wider and faster sampling ~\cite{koch2015mapping}.

To mitigate this limitation, our proposed automated ICM prototype is a small-scale, lightweight ($1040$\;g), and scalable device that can get airborne using accessible and commercially available aerial robots. The ICM employs a novel trapping structure that provides reliable insect capture and a survivable internal environment during aerial transportation. 
The device integrates a custom designed Ultraviolet (UV) LED core as a visual attractant for insects. UV light is a broad attractant that has long been used for management of houseflies in indoor, urban, and agricultural situations~\cite{hogsette2008ultraviolet}. The UV light can be accurately and precisely controlled by an onboard computer, which is needed for automation and is a key benefit of our ICM. 
The module is further capable of monitoring and logging environmental temperature, humidity, and insect activities within the device. Development and testing is performed primarily with houseflies (\emph{Musca domestica}). However, in preliminary tests the ICM is found capable of capturing and retaining mosquitoes (\emph{Aedes aegypti}) as well. Its modular design features could enable adaptation for other insect species too. 

Our proposed device may also find application in tracking ecological changes. Periodically sampling insects in restoring wetland environments could help study biodiversity changes over time~\cite{mayton2017networked}. Information extracted from sampled insects might provide a deeper insight into the effects of rising global temperatures on the breeding of insects and spreading of vector-borne disease in warming climates~\cite{wu2016impact}.

\section{Key Components of the Insect Capture Module and its Design}\label{sec:MaterialsMethods}

\begin{figure}[!h]
	\vspace{-3pt}
	\centering
\includegraphics[trim={0cm 0cm 13cm 2cm},clip,width=\linewidth]{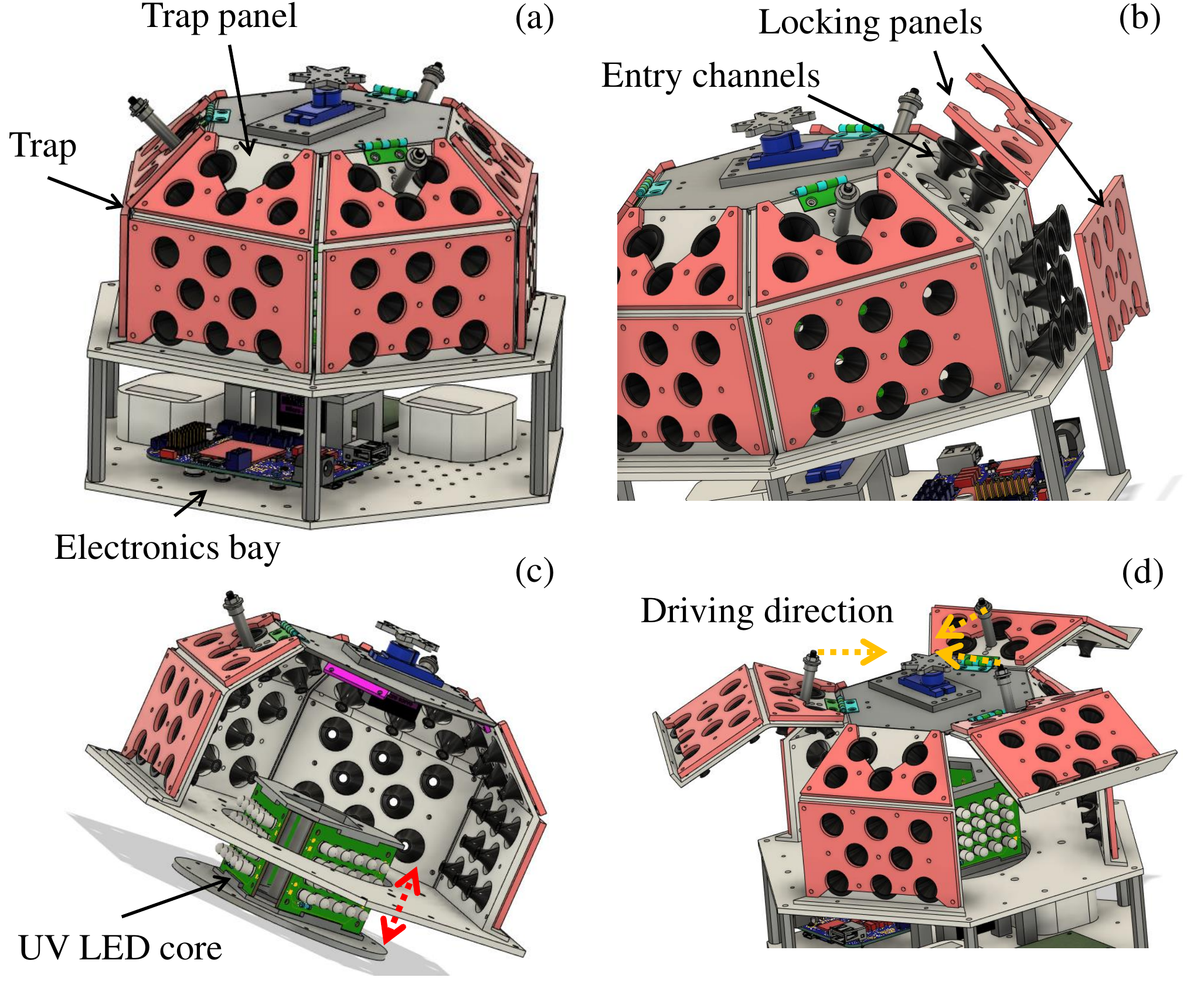}
	\vspace{-0pt}
	\caption{Insect capture module design: (a) Overview of the design; (b) ICM side panel assembly; (c) UV LED core assembly; and (d) ICM side panel actuation mechanism.}
	\label{fig:trap_design}
	\vspace{-9pt}
\end{figure}

Our ICM features a UV light attractant, a passive capture mechanism, and panels which can open and close (Fig. \ref{fig:trap_design}). The ICM is integrated with an additional onboard computer enabling autonomous control or teleoperation via a WiFi connection. Additional sensors such as a digital thermometer and humidity sensor are used to monitor the environment. A small camera is used to observe insect activity inside ICM.




\subsection{Insect Capture Module}
The design of our ICM is driven by the need to transport and deploy ICM on an aerial robot, capture insects efficiently, and keep the insects alive during transportation between the target site and the laboratory. ICM is designed so that a user can easily mount or dismount the device from an aerial robot for experiment preparation or maintenance. 

The module consists of the insect trap and electronics bay (Fig.~\ref{fig:trap_design}(a)). The visual lure within the trap is a custom-made UV LED core with no moving parts, and is embedded within a compact, light-weight, and dome-shaped frame. The dome shape aims to reduce the effects of drag induced by ICM when airborne, which may help offset the increased energy consumption of the aerial robot due to the additional payload.

The six side panels of ICM serve the dual purpose of capturing insects in the environment as well as providing structural integrity to hold the servo motors and cable mechanism to actuate the panels (Fig.~\ref{fig:trap_design}(d)). The design is modular so that panels can be replaced easily if one is damaged; each panel contains 13 insect entry ports. Figure~\ref{fig:trap_design}(b) illustrates the locking panels which can be removed to change the entry channels. The insect capture mechanism can be accessed directly from the bottom of the base plate (Fig.~\ref{fig:trap_design}(c)). 

The total weight of the assembled ICM is $1040$\;g (including two temperature/humidity sensors that weigh $200$\;g, and one $3000$\;mAh LiPo battery that weighs $200$\;g); the device measures $217$\;mm L x $188$\;mm W x $162$\;mm H. Most mechanical parts were 3D printed (Markforged Mark Two and MakerBot Replicator+ 3D printers). Larger parts with planar surfaces such as the base and electronics bay plates were laser-cut (Universal Laser Systems VLS3.60 laser cutter; $1/8$\;in. nylon sheets). The whole device can be assembled and disassembled with simple hand tools easily to facilitate post-experiment analysis or cleaning for next use.


\subsection{Entry Channel Design}
The ICM entry channels are funnel-shaped (Fig.~\ref{fig:trap_design}(b)) to facilitate insects' entry but hinder their escape. A trade-off should be noted that as the inward diameter decreases to prevent insects from escaping, but decrease in diameter might also discourage insects from entering the ICM. In Section~\ref{sec:entry_channel_diameter} we discuss the experiments performed that helped us determine $4$\;mm as an optimal inward funnel diameter to allow houseflies entering and keeping them trapped inside. 


\subsection{Light Source Insect Attractant}

\begin{figure}[!ht]
	\vspace{6pt}
	\centering
\includegraphics[trim={0 0 0 0},clip,width=0.65\linewidth]{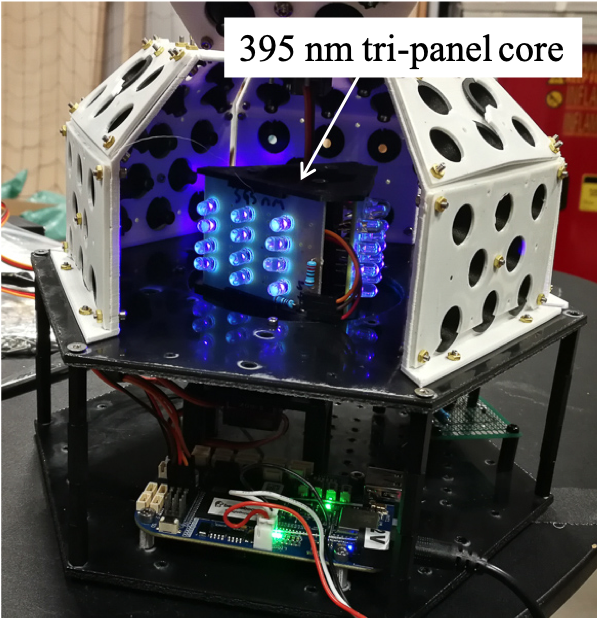}
	\vspace{0pt}
	\caption{Close-up view of ICM with our custom UV LED core (a tri-panel configuration) and BeagleBone Blue onboard computer.}
	\label{fig:trap_electronics}
	\vspace{0pt}
\end{figure}

We designed and manufactured custom printed circuit board (PCB) panels to test different UV light wavelengths and light emitting diode (LED) arrangements. The primary PCB panel consists of a 4x3 UV LED array and MOSFET circuit that allows the general purpose pins (GPIOs) of the onboard computer (BeagleBone Blue) to control the panels to be in either an off, on, or flickering state. 
All PCB panels are attached together using a 3D printed triangular prism shaped mount. We considered single-panel and tri-panel configurations. The single-panel configuration emits light over a third of the ICM's interior. The tri-panel configuration (Fig. \ref{fig:trap_electronics}) emits UV light on all interior sides of the ICM. We have also used UV LED strips, instead of UV LED arrays, to attach to the 3D printed core.

Behavioral studies in the literature suggest that UV light between $300-400$\;nm is the most attractive to both male and female houseflies \cite{green2017role,smallegange2003attractiveness}. We have used UV wavelengths between $365-395$\;nm given the availability of commercial off-the-shelf UV LED parts. All UV LED core configurations tested are listed in Table \ref{table:led_configurations}. 

\begin{table}[!h]
\vspace{0pt}
\caption{Ultraviolet (UV) LED Core Configurations.}
\vspace{-9pt}
\label{table:led_configurations}
\begin{center}
\begin{tabular}{llr}
\toprule
Wavelength (nm)    & Type & Part \\
\midrule
395      & LED strip     & Waveform Lighting, realUV \\
         & Single-panel  & BIVAR, UV5TZ-395-30\\
         & Tri-panel     & BIVAR, UV5TZ-395-30\\
         \midrule
385      & Single-panel  & VCC, VAOL-5GUV8T4\\
         & Tri-panel     & VCC, VAOL-5GUV8T4\\
         \midrule
365      & LED strip     & Waveform Lighting, realUV\\
\bottomrule
\end{tabular}
\end{center}
\vspace{-16pt}
\end{table}

\subsection{Electronics}
The BeagleBone Blue (BBBL) is a small Linux-based computer and is used to control the UV LED core and ICM panels. The BBBL supports remote access via a WiFi connection either through the BBBL access point or by connecting the BBBL to an external WiFi network. The BBBL GPIOs can output logic HIGH/LOW signals which are used to enable the logic level MOSFET circuit, located on each LED core PCB panel, that controls the LEDs. The BBBL also has a servo driving circuit which is used to control the servo motors that actuate the ICM's side panels.

A 2S LiPo battery powers the BBBL and servo motors directly, and a buck converter voltage regulator is connected to the battery to power the LED core. A mini 1080p camera with night vision is placed inside the ICM to record insect activity. Digital thermometers record the temperature inside and outside the ICM, and a humidity sensor records the humidity outside of the ICM. The power supply, BBBL, and temperature/humidity sensors are housed in the electronics bay as shown in Fig. \ref{fig:trap_design}(a).

\section{Experimental Methods and Results}\label{sec:Experiment}

We performed four sets of experiments to assess individual components of the ICM and its performance in near real-world testing conditions. Unless otherwise stated, all experiments were performed with either houseflies (\textit{Musca domestica}), mosquitoes (\textit{Aedes aegypti}), or a combination of both. 
We used the DJI Matrice 100 quadcopter as the aerial robot platform base (see Fig.~\ref{fig:flyability}). The DJI M100 has a maximum takeoff weight of $3600$\;g and can hover for $16$\;min with $1000$\;g payload (using a single TB48D battery).

\subsection{Entry Channel Diameter}\label{sec:entry_channel_diameter}
\textbf{Objective:}
To determine suitable design parameters for an entry channel capable of capturing houseflies and retaining them inside the ICM. 

\begin{figure}[!h]
	\vspace{-3pt}
	\centering
\includegraphics[trim={0 0 0 0},clip,width=0.7\linewidth]{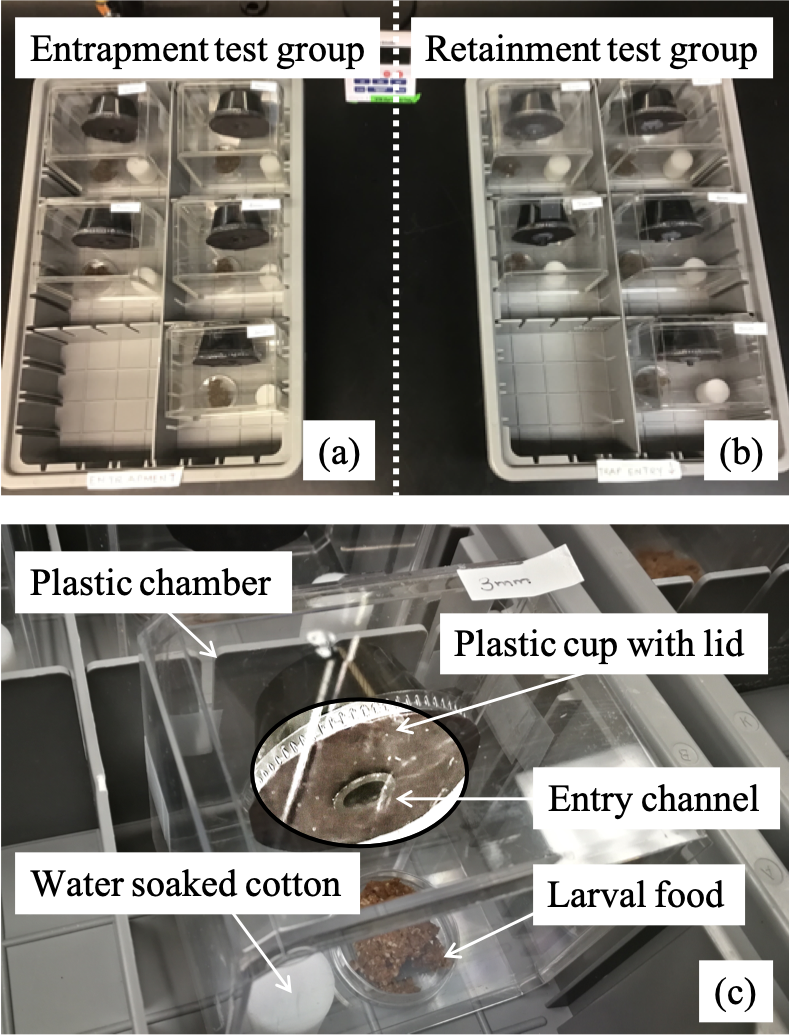}
	\vspace{0pt}
	\caption{(a) Entrapment test group and (b) retainment test group. (c) Detailed chamber setup.}
	\label{fig:entry_channel_test_setup}
	\vspace{-6pt}
\end{figure}

\textbf{Setup:}
We designed funnel-shaped entry channels, with a $15$\;mm wide outer opening and varying inner opening diameter ranging from $3$\;mm to $7$\;mm such that flies enter the ICM and do not come out. We performed mock insect `\emph{entry/entrapment}’ and `\emph{retainment}’ tests (Fig. \ref{fig:entry_channel_test_setup}) with groups of houseflies (5 flies per test) for each entry channel size in order to determine the right channel size for the ICM. The \emph{entrapment} test is designed to assess insect entering effectiveness, and the \emph{retainment} test is designed to assess insect retaining effectiveness.


Each entrapment test was conducted in a transparent plastic chamber. Inside the chamber we placed a small black plastic cup containing $5$ houseflies. An entry channel was affixed onto the cup's lid such that the larger outward side is directed toward the inside of the cup. This configuration is useful in determining whether the flies located inside the cup will move through the entry channel and into the chamber. Flies were motivated to leave the cup using an insect attractant. The insect attractant consisted of larval food and cotton soaked in water located inside the chamber, together with the room's fluorescent lighting. The experiment's duration was $24$ hrs. All five inner diameter funnel configurations ($3$\;mm, $4$\;mm, $5$\;mm, $6$\;mm, $7$\;mm) were tested simultaneously for each trial. 

Retainment tests were conducted using the same setup, but by reversing the orientation of the entry channel so that the smaller inward side was toward the inside of the cup.

\begin{figure}[!h]
	\vspace{6pt}
	\centering
\includegraphics[trim={0 0 0 0},clip,width=1.0\linewidth]{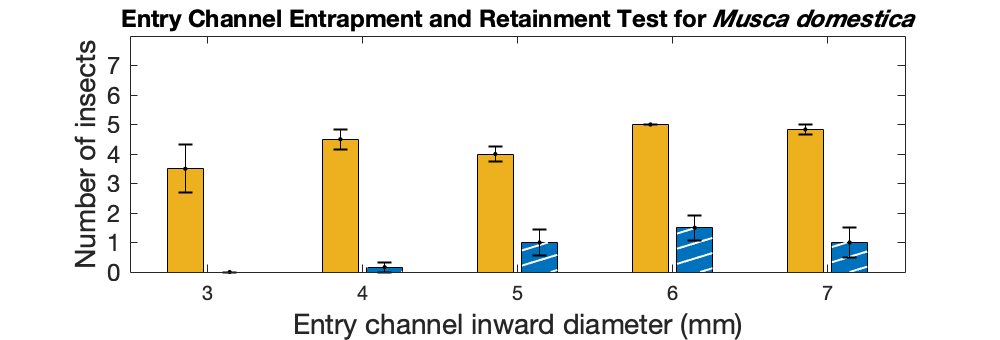}
	\vspace{-12pt}
	\caption{Number of houseflies inside the chamber affixed with different sized entry channels after $24$ hrs. (solid) Number of insects that entered the chamber, and (hatched) number of insects that escaped the chamber. $n=6$ trials, 25 flies/trial, 5 flies/channel diameter.}
	\label{fig:entry_channel_chart}
	\vspace{0pt}
\end{figure}

\textbf{Results:}
All entry channel tests were performed for $6$ trials under the same environment conditions. The number of houseflies inside the black cup were counted after every trial. Results are shown in Fig.~\ref{fig:entry_channel_chart}. We observe that larger inner diameters ($6-7$\;mm) allow more flies to enter the chamber. However, they cannot retain them well. On the other hand, the smallest passage ($3$\;mm) retains flies very well, if they get in the chamber (notice the high variability across trials). 

Overall trap performance should be determined by the number of flies both entering and staying in the chamber. From Fig.~\ref{fig:entry_channel_chart} we can compute the entrapping rate $R_{entrap}$ and retaining rate $R_{retain}$. 
To measure overall effectiveness, we further define 
the \emph{capturing} rate $R_{capture}$ by 
\begin{equation}
    R_{capture} = R_{entrap} \cdot R_{retain}
\end{equation}

The results presented in Table \ref{table:entry_channel_capture_rate} show that a $4$ mm channel has the highest capturing rate and therefore, is selected for use in our ICM in further experiments.

\begin{table}[!h]
\vspace{-3pt}
    \caption{Entry Channel Capturing Performance Analysis for Houseflies (\emph{Musca domestica}). $n=6$ trials, 25 flies/trial, 5 flies/channel diameter.}
    \vspace{-6pt}
    \label{table:entry_channel_capture_rate}
    \begin{center}
    \begin{tabular}{|c|c|c|c|c|c|}
    \hline
    Inward diameter & 3 mm & \textbf{4 mm} & 5 mm & 6 mm & 7 mm\\
    \hline
    Entrapping rate(\%) & $70.0$ & $90.0$ & $80.0$ & $100$ & $96.7$\\
    \hline
    Retaining rate(\%) & $100$ & $96.7$ & $80.0$ & $70.0$ & $80.0$\\
    \hline
    Capture rate(\%) & $70.0$ & $\mathbf{87.0}$ & $64.0$ & $70.0$ & $77.3$\\
    \hline
    \end{tabular}
    \end{center}
    \vspace{-6pt}
\end{table}

It should also be noted that all houseflies inside the chamber were alive after each 24-hr experiment, provided there was food and water available.

\subsection{Insect Capture with Ultraviolet Light}\label{sec:insect_capture_UV_light}

\textbf{Objective:} To determine a feasible UV LED wavelength range and LED core configuration for capturing insects in lighted (maximum room light) and dark (no room light) environments. 

\textbf{Setup:} Similar experiments were performed on houseflies and mosquitoes. 


\begin{figure}[!h]
\vspace{-6pt}
      \centering
      \begin{subfigure}{0.22 \textwidth}
        \includegraphics[trim={0 0 0 0}, clip, width=\textwidth]{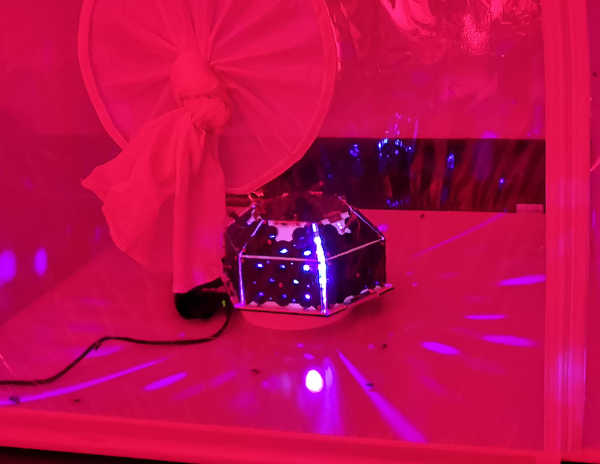}
      \end{subfigure}
      \begin{subfigure}{0.22 \textwidth}
        \includegraphics[trim={0 0 0 0}, clip, width=\textwidth]{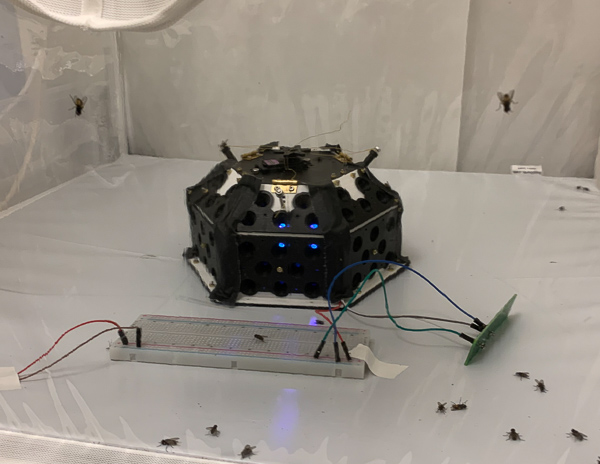}
      \end{subfigure}
      \caption{Wavelength test in a dark environment with red light (left) and a regularly-lit environment (right).}
      \label{fig:wavelength_test}
	\vspace{-9pt}
\end{figure}

\textit{Houseflies:} An ICM with a UV LED core was placed inside a $60$\;x\;$60$\;x\;$60$\;cm cage. A group of $50$ houseflies (a mix of males and females, $5-10$ days old) were allowed to acclimate in the room for $30$\;min and then gently released into the cage. The UV LED core was turned on and remained so for two hours. At the end of the $2$-hr capture period, the UV LED core was turned off and all flies in the cage were captured using a hand-held mechanical aspirator. Then, the ICM was transferred from the cage to a second empty observation cage. The ICM side panels were opened to release any flies that were trapped. Flies in the observation cage were captured with another aspirator and stored in test tubes to be knocked down in a freezer and counted later on. 

\begin{figure}[!h]
	\vspace{-6pt}
	\centering
\includegraphics[trim={0 0 0 0},clip,width=1.0\linewidth]{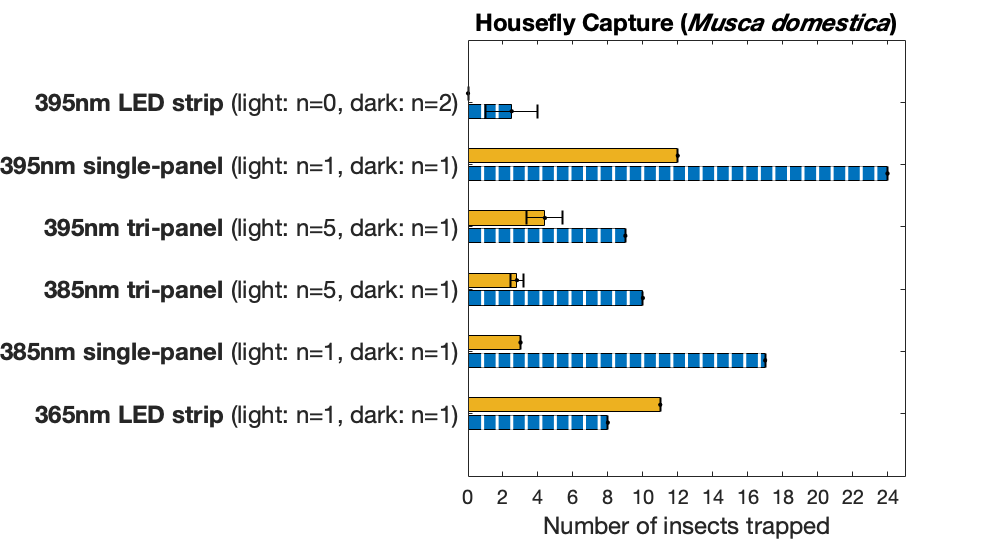}
	\vspace{-6pt}
	\caption{Number of houseflies trapped (out of $50$) after $2$ hrs in cage with ICM affixed with LED cores of different wavelengths in regular (solid bars) and low (hatched bars) ambient lighting conditions, respectively. Parameter $n$ refers to the number of trials, 50 flies/trial.}
	\label{fig:wavelength_test_flies_chart}
	\vspace{-6pt}
\end{figure}

The procedure was repeated for each of the LED core configurations listed in Fig.~\ref{fig:wavelength_test_flies_chart} in both lighted and dark environments. In the first case, the room light was set to regular brightness for the $30$\;min acclimation period and the $2$-hr capture period. For a dark environment, the light was off for the $30$\;min acclimation and the $2$-hr capture periods.

\begin{figure}[!h]
	\vspace{6pt}
	\centering
	\includegraphics[trim={0 0cm 0 0},clip,width=1.0\linewidth]{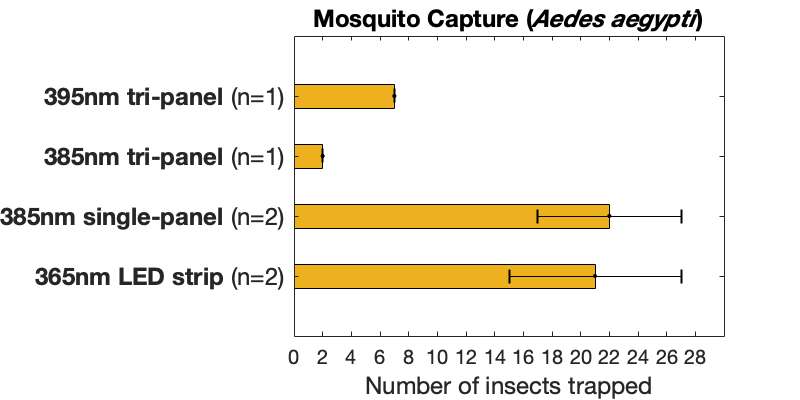}
	\vspace{-12pt}
	\caption{Number of mosquitoes trapped (out of $50$) after $2$ hrs in a regularly-lit environment using the ICM affixed with LED cores of different wavelengths. Parameter $n$ refers to the number of trials, 50 flies/trial.}
	\label{fig:wavelength_test_mosquitoes_chart}
	\vspace{-3pt}
\end{figure}

\textit{Mosquitoes:} The procedure for mosquitoes is nearly identical to houseflies. A group of $50$ female \textit{Aedes aegypti} mosquitoes ($5-12$ days old) were released into the $60$\;x\;$60$\;x\;$60$\;cm cage with the ICM. Insect capture occurred for $2$-hr at the end of which the ICM was moved to an empty observation cage and left overnight. Any mosquitoes previously trapped that had escaped the ICM into the observation cage by the following morning were recorded. This procedure is repeated using the UV LED core configurations listed in Fig.~\ref{fig:wavelength_test_mosquitoes_chart}.

\textbf{Results:} 
The number of insects inside the ICM were recorded after every trial. The results are shown in Fig. \ref{fig:wavelength_test_flies_chart} for houseflies and Fig. \ref{fig:wavelength_test_mosquitoes_chart} for mosquitoes. 

Results of testing with houseflies (Fig.~\ref{fig:wavelength_test_flies_chart}) indicate the feasibility of the ICM to capture houseflies in a controlled environment. Despite the low number of trials, results suggest that for all cases the device can yield higher capture rates at low ambient light conditions, which is reasonable to expect as the UV light source is more concentrated and hence increases visual contrast with its surrounding and thus becomes a stronger attractant. 
By looking at tri-panel core configurations, results suggest that $395$\;nm UV LEDs might be more appropriate than $385$\;nm UV LEDs, at least under regular ambient light conditions. 
A single-panel $395$\;nm UV LED core configuration appears to perform better than others; however, more data are needed to confirm this finding. 

It is worth highlighting that whilst entry channel and UV LED core configurations of the trap are based on houseflies, we also performed experiments with mosquitoes. Testing the performance of the device with a different insect species can offer preliminary indication on its capacity to attract other insects which in turn may offer an additional pool of data to study presence of pathogens at an area of interest. Our preliminary results shown in Fig.~\ref{fig:wavelength_test_mosquitoes_chart} confirm that the trap can capture mosquitoes although it was originally designed to capture houseflies. Results suggest that the tri-panel core might not perform as well as a single-panel core or a core fitted with an LED strip. UV LED wavelength may not play an as fundamental role as in houseflies for attracting mosquitoes. (More data are needed to confirm these observations.)

Further, an important observation was made when comparing the amount of heat dissipated from a core made of LED strips versus a panel made of an LED array. The LED strips dissipated much more heat which greatly warmed the ICM over the 2-hr capture period. The LED arrays did not produce nearly as much heat, but they were much lower in brightness than the LED strips. 

\subsection{Transportation Reliability Test}

\textbf{Objective 1:} To assess the performance of a quadcopter with mounted ICM during take-off, flight, and landing in a preprogrammed flight path.

\textbf{Objective 2:} To assess the survival, mortality, and escaping of houseflies contained within the ICM during aerial transportation.

\begin{figure}[h]
	\vspace{6pt}
	\centering
\includegraphics[trim={0cm 0cm 0cm 0cm},clip,width=1.0\linewidth]{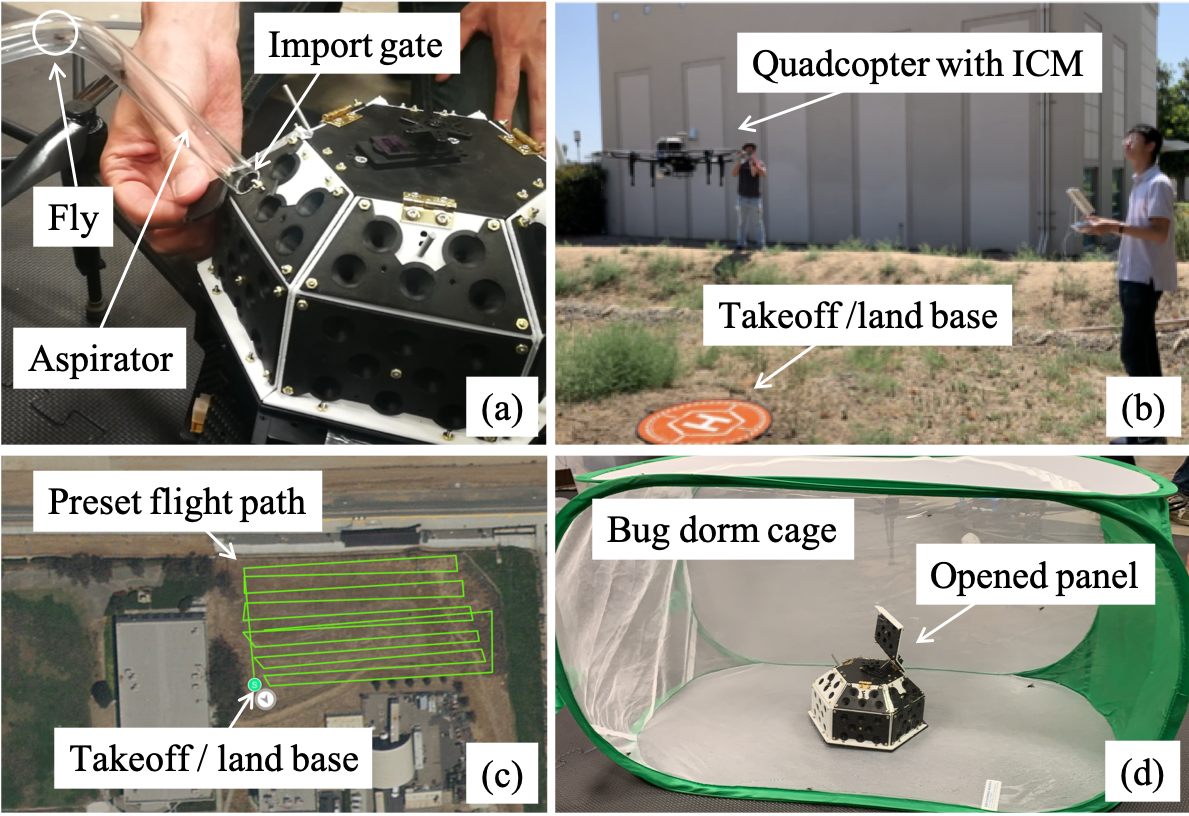}
	\vspace{-9pt}
	\caption{Transportation reliability test procedure: (a) Import flies. (b) Quadcopter takes off. (c) Track path and land. (d) Dismount trap and retrieve flies.}
	\label{fig:flyability_procedure}
	\vspace{-6pt}
\end{figure}

\textbf{Setup:} Houseflies were transported to the testing station in a cooled chamber provided with humidity and 10\% sugar water. A total of $50$ houseflies were then individually transferred using an aspirator inside the ICM via a modified port with a swivel locking cover (Fig.~\ref{fig:flyability_procedure}(a)). The ICM was fitted with the $4$\;mm size entry channels which were demonstrated to have the most effective performance in capturing houseflies as described in Section~\ref{sec:entry_channel_diameter}. The ICM was then mounted onto the quadcopter and the digital sensors started to record the temperature and humidity. The quadcopter and ICM were then taken outside and placed in an open field (Fig. \ref{fig:flyability_procedure}(b)). 

The quadcopter was commanded to fly preset paths with a forward velocity of $5$\;m/s and $2$\;m/s for $5$ min at an altitude of $30$\;m (Fig.~\ref{fig:flyability_procedure}(c)). At the end of each flight session, the ICM was dismounted from the aerial robot and placed in a large cage. The ICM panels were opened to release the houseflies into the cage (Fig.~\ref{fig:flyability_procedure}(d)). Houseflies in the cage were captured with an insect aspirator and stored in test tubes to be knocked down in a freezer and counted later on.

\begin{figure}[!h]
	\vspace{0pt}
	\centering
\includegraphics[trim={5cm 0cm 5cm 0cm},clip,width=1\linewidth]{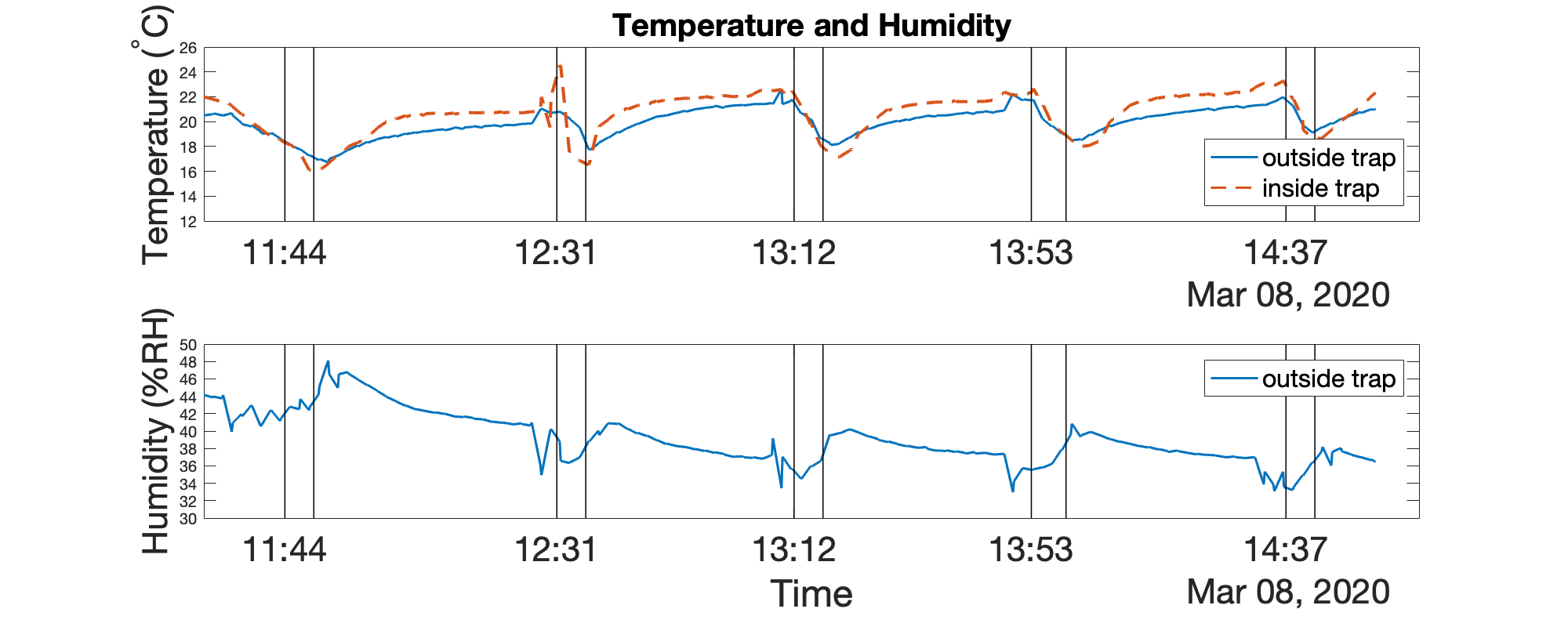}
	\vspace{-9pt}
	\caption{Transportation reliability test: Temperature and humidity measured over $5$ trials. The black vertical lines indicate the start and end of a $5$-min experiment.}
	\label{fig:Flyability_temperature_humidity}
	\vspace{-9pt}
\end{figure}

\textbf{Results:} We counted the number of houseflies that remained alive over $6$ trials for a quadcopter flying at $5$\;m/s. Out of a total of $50$ houseflies, on average $49$ of them were alive. During each flight experiment, a general decrease in the temperature inside/outside the ICM and increase in relative humidity outside the ICM was observed (Fig. \ref{fig:Flyability_temperature_humidity}). The same experiment was repeated for the robot flying with a $2$\;m/s forward velocity; $47$ out of $50$ flies remained alive by the end of the flight; three flies had escaped during the counting process at the end.

In these sets of experiments we demonstrated that it is feasible to use the current ICM design to transport houseflies alive at low-to-medium cruise speeds. Houseflies can survive inside the ICM environment under variable turbulent air flow. The ICM has potential for transporting larger batch sizes; however, alterations in the ICM's design may be needed to prevent insects from escaping, due to increased density.

Further, we flew the quadcopter retrofitted with the ICM but without houseflies loaded at forward velocities of up to $10$\;m/s to assess the performance of the quadcopter with ICM during high speed flight. We did not observe any structural or electronics damage to the ICM during the tests. Flying the quadcopter at $10$\;m/s would be sufficient for future experiments where large areas would need to be covered.    

\subsection{Automatic Remote Insect Collection}
\textbf{Objective:} To perform semi-field tests by emulating insect capture and transport using an aerial robot equipped with ICM. 

\textbf{Setup:} Tests were performed in a $6.4$\;m L x $3.2$\;m W x $2.2$\;m H greenhouse with screened mesh walls lined with plastic tarp blocking lighting from adjacent units. The chamber is conditioned to a set temperature earlier in the day by using swamp coolers, space heater and fans. Over $100$ houseflies (a mix of male and female, each $5-10$ days old) were held in a cage and allowed to acclimate in the greenhouse for at least $2$\;hrs. 

\begin{figure}[!t]
	\vspace{6pt}
	\centering
\includegraphics[trim={0 0.3cm 0 2cm},clip,width=0.8\linewidth]{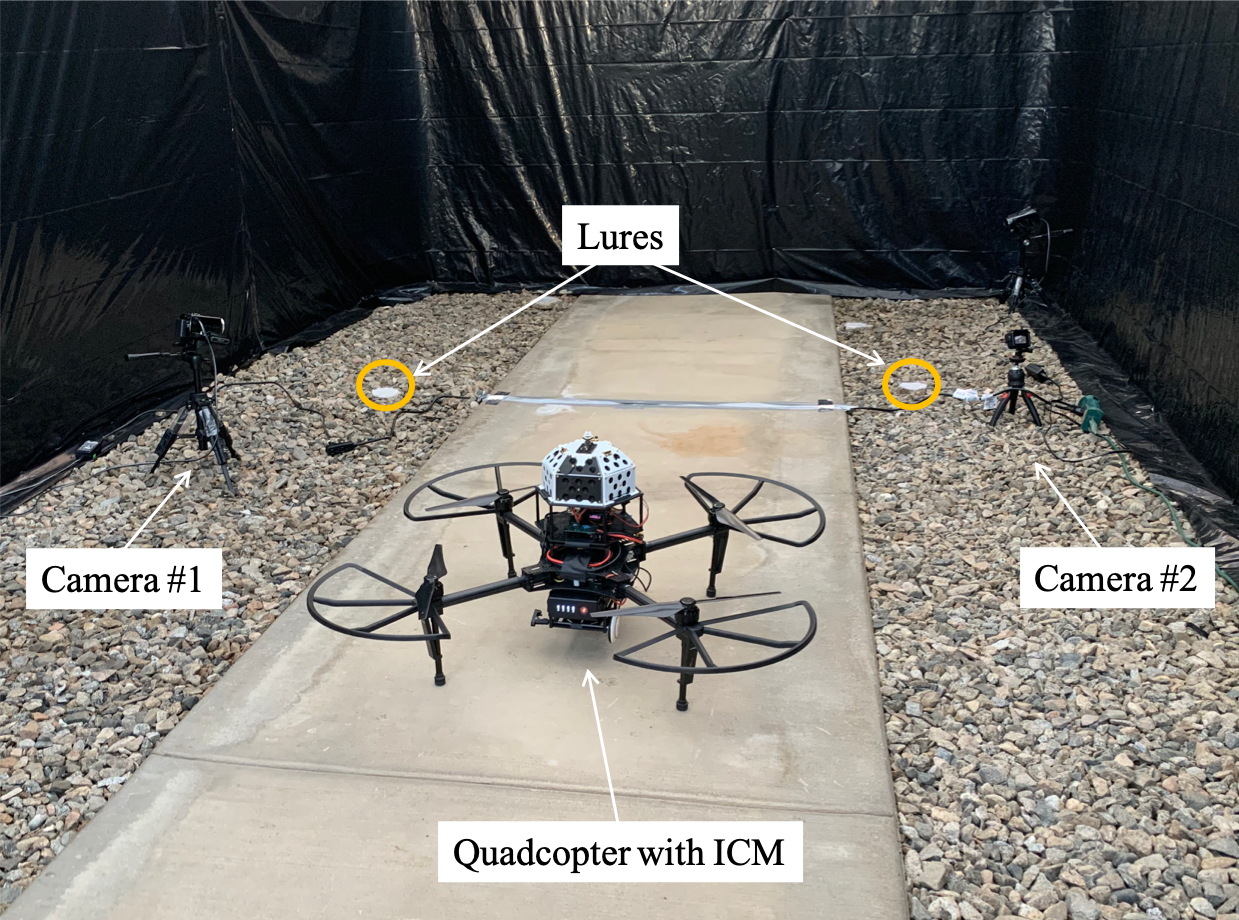}
	\vspace{0pt}
	\caption{Automatic remote insect sampling test: A quadcopter equipped with ICM to perform insect sampling. An operator located outside of the room remotely operates the trap.}
	\label{fig:darpa_greenhouse}
	\vspace{-16pt}
\end{figure}

The aerial robot with mounted ICM was placed in the center of the room (Fig.~\ref{fig:darpa_greenhouse}). Lures were placed inside the chamber to emulate external influences for insect capturing. Cameras were setup to record fly activity. Houseflies were released into the room and the UV LED core was remotely turned on. The ICM remained in capture mode with the UV LEDS on for $2$\;hrs. At the end of the capture period, the aerial robot was flown to hover in place for $3$\;min to emulate insect transport and turbulent air conditions. $20$\% of the battery level was allocated for flight to the target site, $50$\% to keep the quadcopter on during insect capture, and $20$\% to fly the quadcopter back. After the aerial robot landed, the ICM was dismounted from the aerial robot, sealed in a plastic container, and transferred to the lab to examine contents. Captured houseflies were frozen and counted. 

\textbf{Results:}
This experiment was performed twice over two days under similar environmental conditions. The ICM captured $7$ houseflies during the first trial and $5$ during the second. Despite the external influence of the lures, the ICM was able to capture flies within a $2$-hr time-span. 

\section{Discussion and Outlook}

The paper demonstrated the feasibility of a novel airborne approach for remote collection, transfer, and analysis of insects. Insects often transmit different pathogens to humans. Rapid and large-scale collection of insects at specified geographical locations can enable a more robust and timely prediction of vector-borne disease outbreaks and spread, significantly aiding public health officials and researchers to better understand the dynamics of the evolution of a disease.

As an important step in the development and feasibility testing is the semi-field trial of the novel automated insect capture module which we designed and fabricated in house. The automated ICM was found capable of luring, trapping and safely transporting houseflies (\textit{Musca domestica}). The developed ICM is a small-scale, lightweight and relatively low cost device that can be manufactured through benchtop prototyping tools like laser cutting and 3D printing. Taken together, these features make the device accessible to use by researchers. The device is also compatible with commercially available aerial robots so it can be deployed in remote places. ICM is modular and features a custom-made UV LED core as the means to attract insects. The insect sampling scheme was successfully demonstrated in both laboratory and semi-field tests. In addition, preliminary testing indicated that the ICM prototype may be also able to capture and retain other insects like mosquitoes as well. This versatility can be extremely important in practice since it could give researchers the ability to conduct surveillance of multiple insect species in an area without the need to redeploy the insect trap.

Despite its effectiveness in the feasibility testing presented herein, further investigation is necessary to optimize the ICM performance. Beyond testing with larger sample sizes, the evaluation of various lures that are species specific, insect survivability under faster flight, and total energy efficiency during remote field operations are important to test. 

Overall, our ICM shows promise for fast, efficient, and adaptive sample collection. Integration of more advanced chemical/genomic analysis methodologies downstream and more intelligent robotic techniques offers opportunities to facilitate scalable environmental monitoring and improve prediction of large-scale disease trending and outbreaks.

\balance
\bibliographystyle{IEEEtran}
\bibliography{citation.bib}

\begin{thebibliography}{10}
\providecommand{\url}[1]{#1}
\csname url@samestyle\endcsname
\providecommand{\newblock}{\relax}
\providecommand{\bibinfo}[2]{#2}
\providecommand{\BIBentrySTDinterwordspacing}{\spaceskip=0pt\relax}
\providecommand{\BIBentryALTinterwordstretchfactor}{4}
\providecommand{\BIBentryALTinterwordspacing}{\spaceskip=\fontdimen2\font plus
\BIBentryALTinterwordstretchfactor\fontdimen3\font minus
  \fontdimen4\font\relax}
\providecommand{\BIBforeignlanguage}[2]{{%
\expandafter\ifx\csname l@#1\endcsname\relax
\typeout{** WARNING: IEEEtran.bst: No hyphenation pattern has been}%
\typeout{** loaded for the language `#1'. Using the pattern for}%
\typeout{** the default language instead.}%
\else
\language=\csname l@#1\endcsname
\fi
#2}}
\providecommand{\BIBdecl}{\relax}
\BIBdecl

\bibitem{Yangeabb5589}
G.-Z. Yang, B.~J.~Nelson, R.~R. Murphy, H.~Choset, H.~Christensen,
  S.~H.~Collins, P.~Dario, K.~Goldberg, K.~Ikuta, N.~Jacobstein, D.~Kragic,
  R.~H. Taylor, and M.~McNutt, ``Combating covid-19{\textemdash}the role of
  robotics in managing public health and infectious diseases,'' \emph{Science
  Robotics}, vol.~5, no.~40, 2020.

\bibitem{tirado2019drones}
F.~Tirado and P.~T. Cano, ``Drones and epidemiology: A new anatomy for
  surveillance,'' \emph{BioSocieties}, pp. 1--19, 2019.

\bibitem{delmerico2019current}
J.~Delmerico, S.~Mintchev, A.~Giusti, B.~Gromov, K.~Melo, T.~Horvat, C.~Cadena,
  M.~Hutter, A.~Ijspeert, D.~Floreano, L.~M. Gambardella, R.~Siegwart, and
  D.~Scaramuzza, ``The current state and future outlook of rescue robotics,''
  \emph{Journal of Field Robotics}, vol.~36, no.~7, pp. 1171--1191, 2019.

\bibitem{das2011towards}
J.~{Das}, T.~{Maughan}, M.~{McCann}, M.~{Godin}, T.~{O'Reilly}, M.~{Messié},
  F.~{Bahr}, K.~{Gomes}, F.~{Py}, J.~G. {Bellingham}, G.~S. {Sukhatme}, and
  K.~{Rajan}, ``Towards mixed-initiative, multi-robot field experiments:
  Design, deployment, and lessons learned,'' in \emph{IEEE/RSJ International
  Conference on Intelligent Robots and Systems (IROS)}, 2011, pp. 3132--3139.

\bibitem{qian2017ground}
F.~Qian, D.~Jerolmack, N.~Lancaster, G.~Nikolich, P.~Reverdy, S.~Roberts,
  T.~Shipley, R.~S. Van~Pelt, T.~M. Zobeck, and D.~E. Koditschek, ``Ground
  robotic measurement of aeolian processes,'' \emph{Aeolian research}, vol.~27,
  pp. 1--11, 2017.

\bibitem{howard2006experiments}
A.~Howard, L.~E. Parker, and G.~S. Sukhatme, ``Experiments with a large
  heterogeneous mobile robot team: Exploration, mapping, deployment and
  detection,'' \emph{The International Journal of Robotics Research}, vol.~25,
  no. 5-6, pp. 431--447, 2006.

\bibitem{sarwar2015insect}
M.~Sarwar, ``Insect vectors involving in mechanical transmission of human
  pathogens for serious diseases,'' \emph{International Journal of
  Bioinformatics and Biomedical Engineering}, vol.~1, no.~3, pp. 300--306,
  2015.

\bibitem{khamesipour2018systematic}
F.~Khamesipour, K.~B. Lankarani, B.~Honarvar, and T.~E. Kwenti, ``A systematic
  review of human pathogens carried by the housefly (musca domestica l.),''
  \emph{BMC public health}, vol.~18, no.~1, p. 1049, 2018.

\bibitem{koch2015mapping}
T.~Koch, ``Mapping medical disasters: Ebola makes old lessons, new,''
  \emph{Disaster medicine and public health preparedness}, vol.~9, no.~1, pp.
  66--73, 2015.

\bibitem{karydis2017energetics}
K.~Karydis and V.~Kumar, ``Energetics in robotic flight at small scales,''
  \emph{Interface Focus}, vol.~7, no.~1, p. 20160088, 2017.

\bibitem{pinniger1990food}
D.~Pinniger, ``Food-baited traps; past, present and future,'' \emph{Journal of
  the Kansas Entomological Society}, pp. 533--538, 1990.

\bibitem{mcneil1991behavioral}
J.~N. McNeil, ``Behavioral ecology of pheromone-mediated communication in moths
  and its importance in the use of pheromone traps,'' \emph{Annual Review of
  Entomology}, vol.~36, no.~1, pp. 407--430, 1991.

\bibitem{barghini2012uv}
A.~Barghini and B.~A. Souza~de Medeiros, ``{UV} radiation as an attractor for
  insects,'' \emph{Leukos}, vol.~9, no.~1, pp. 47--56, 2012.

\bibitem{schneider1982flying}
W.~A. Schneider, ``Flying insect trap,'' Jun.~1 1982, {US Patent} 4,332,100.

\bibitem{santos2018fan}
D.~A. Santos, R.~F. Brandao, G.~A. Duarte, D.~R. Totti, V.~Furtado, and J.~J.
  Rodrigues, ``A fan-based smart selective trap for flying insects,'' in
  \emph{IEEE 10th Latin-American Conference on Communications (LATINCOM)},
  2018, pp. 1--5.

\bibitem{miller2001counterflow}
M.~H. Miller, B.~E. Wigton, and K.~Lonngren, ``Counterflow insect trap,''
  Sep.~11 2001, {US Patent} 6,286,249.

\bibitem{shanahan1977baseboard}
F.~V. Shanahan and H.~H. Feller, ``Baseboard trap for crawling insects,''
  Sep.~20 1977, {US Patent} 4,048,747.

\bibitem{linn2016building}
A.~Linn, ``Building a better mosquito trap,'' \emph{International Pest
  Control}, vol.~58, no.~4, p. 213, 2016.

\bibitem{dacko2020mosquito}
N.~M. Dacko, M.~R. Nava, C.~Vitek, and M.~Debboun, ``Mosquito surveillance,''
  in \emph{Mosquitoes, Communities, and Public Health in Texas}.\hskip 1em plus
  0.5em minus 0.4em\relax Elsevier, 2020, pp. 221--247.

\bibitem{ching2017insect}
A.~Ching, E.~Jackson, M.~J. Sinclair, and P.~Therien, ``Insect trap,'' Dec.~14
  2017, {US Patent App.} 15/178,354.

\bibitem{hogsette2008ultraviolet}
J.~Hogsette, ``Ultraviolet light traps: design affects attraction and
  capture.'' in \emph{6th International Conference on Urban Pests}, 2008, pp.
  193--196.

\bibitem{mayton2017networked}
B.~Mayton, G.~Dublon, S.~Russell, E.~F. Lynch, D.~D. Haddad,
  V.~Ramasubramanian, C.~Duhart, G.~Davenport, and J.~A. Paradiso, ``The
  networked sensory landscape: Capturing and experiencing ecological change
  across scales,'' \emph{Presence: Teleoperators and Virtual Environments},
  vol.~26, no.~2, pp. 182--209, 2017.

\bibitem{wu2016impact}
X.~Wu, Y.~Lu, S.~Zhou, L.~Chen, and B.~Xu, ``Impact of climate change on human
  infectious diseases: Empirical evidence and human adaptation,''
  \emph{Environment international}, vol.~86, pp. 14--23, 2016.

\bibitem{green2017role}
M.~Green, ``Role of {LED} lights in the design of ultra-violet light traps for
  house fly monitoring and control,'' in \emph{9th International Conference on
  Urban Pests}, 2017, pp. 329--333.

\bibitem{smallegange2003attractiveness}
R.~C. Smallegange, \emph{Attractiveness of different light wavelengths, flicker
  frequencies and odours to the housefly (Musca domestica L.)}.\hskip 1em plus
  0.5em minus 0.4em\relax Rijksuniversiteit Groningen, 2003.

\end{thebibliography}
	
	
\end{document}